\documentclass[unnumsec,webpdf,contemporary,large]{oup-authoring-template}%

\graphicspath{{Fig/}}

\theoremstyle{thmstyleone}%
\theoremstyle{thmstyletwo}%
\theoremstyle{thmstylethree}%

\usepackage{graphicx}
\usepackage{amsmath}
\usepackage{multirow}
\usepackage{tabularx}
\usepackage{url}  
\usepackage{algorithm}
\usepackage{algpseudocode}  
\usepackage{natbib}
\usepackage{dsfont}

\begin{document}

\journaltitle{Journal Title Here}
\DOI{DOI HERE}
\copyrightyear{2022}
\pubyear{2019}
\access{Advance Access Publication Date: Day Month Year}
\appnotes{Paper}

\firstpage{1}


\title[MSCoD]{MSCoD: An Enhanced Bayesian Updating Framework with Multi-Scale Information Bottleneck and Cooperative Attention for Structure-Based Drug Design}

\author[1]{Long Xu}
\author[1]{Yongcai Chen}
\author[1]{Fengshuo Liu}
\author[2, $\ast$]{Yuzhong Peng}

\authormark{Xu et al.}
\address[1]{\orgdiv{Guangxi Key Lab of Human-machine Interaction and Intelligent Decision}, \orgname{Nanning Normal University}, \postcode{530001}, \state{Nanning}, \country{China}}
\address[2]{\orgdiv{College of Big Data and Software Engineering}, \orgname{Zhejiang Wanli University}, \postcode{315000}, \state{Ningbo}, \country{China}}


\abstract{\textbf{Motivation:} Structure-Based Drug Design (SBDD) leverages 3D protein structures to create molecules with high binding affinity. A key challenge is modeling complex protein-ligand interactions, which are inherently multi-scale, hierarchical, and asymmetric. Existing methods often fail to capture these properties, limiting their effectiveness.\\\textbf{Results:} To address these limitations, we propose MSCoD, a novel Bayesian updating-based generative framework for structure-based drug design. MSCoD features a Multi-Scale Information Bottleneck (MSIB) for efficient hierarchical feature extraction via semantic compression across multiple abstraction levels. It also includes a multi-head cooperative attention (MHCA) mechanism, which uses asymmetric protein-to-ligand attention to model diverse interaction types and manage the dimensionality gap between proteins and ligands. Comprehensive evaluations demonstrate that MSCoD surpasses state-of-the-art methods on molecular generation benchmarks. Its real-world applicability is confirmed by case studies on difficult targets like KRAS G12D (7XKJ). Additionally, the MSIB and MHCA modules prove transferable, boosting the performance of GraphDTA on standard drug–target affinity prediction benchmarks (Davis and Kiba).\\\textbf{Availability and implementation:} The code and data underlying this article are freely available at \url{https://github.com/xulong0826/MSCoD}.}


\maketitle

\section{\textbf{1 Introduction}}
Structure-Based Drug Design (SBDD) is a key method in modern computational drug discovery. By analyzing comprehensive three-dimensional protein structures and their binding sites~\citep{wang2022deep,isert2023structure,batool2019structure}, SBDD enables the design of molecules that can precisely bind to target proteins and regulate their biological function. However, protein binding sites show great diversity and dynamic changes across different protein families, making the design process challenging. Although recent advances in structural biology and AI tools like AlphaFold~\citep{jumper2021highly} have greatly enriched available structural information, accurately modeling protein-ligand interactions remains difficult. The vast chemical space and complex spatial relationships further complicate efficient screening and design. Therefore, the development of efficient and intelligent molecular generation tools has become an urgent task in drug design~\citep{tang2024survey}, to meet the demands of complex disease treatments and the pharmaceutical industry.

Recent advances in modeling the geometric structures of biomolecules have motivated a promising direction for SBDD. Several new generative methods have been proposed, which can be broadly categorized into autoregressive models, diffusion models, and Bayesian updating methods. Autoregressive models (ARMs) such as AR~\citep{luo2021AR}, GraphBP~\citep{liu2022Graphbp}, and Pocket2Mol~\citep{peng2022pocket2mol} generate 3D molecules by iteratively adding atoms based on the target binding site. However, ARMs tend to suffer from error accumulation and struggle to find an optimal generation order. As an alternative, diffusion-based methods, including TargetDiff~\citep{guan3d2023}, DecompDiff~\citep{pmlr-v202-guan23a}, and BINDDM~\citep{huang2024binding}, adopt a non-autoregressive approach. They learn to reverse a noise-adding process to generate atom types and positions from a Gaussian prior, using SE(3)-equivariant networks to capture spatial interactions and achieve state-of-the-art performance. More recently, Bayesian updating-based methods like MolCRAFT~\citep{qu2024molcraft} have shown competitive performance by operating in a fully continuous parameter space. Despite their success, these methods still face critical limitations. For instance, while BINDDM captures interactions by extracting key protein-ligand combinatorial substructures, we believe protein-ligand interactions can be modeled with even greater refinement. More broadly, most methods typically process molecular and protein features at a single scale, failing to capture the hierarchical nature of protein-ligand interactions, which range from local atomic contacts to global shape complementarity. Furthermore, they do not deeply model the dynamic spatial interactions and asymmetric adaptation processes that characterize real binding events, leading to suboptimal binding affinity predictions and molecular properties.

Recently, multi-scale algorithms have emerged in molecular representation learning~\citep{zhang2024mvmrl} and have demonstrated significant promise for capturing molecular interactions across various spatial and semantic levels. Concurrently, attention mechanisms have become a standard tool for modeling complex relational patterns in molecular systems. Nevertheless, most current approaches employ relatively simple attention architectures, which are insufficient for capturing the inherent asymmetry and complexity of protein-ligand binding. Therefore, there is an increasing demand for more advanced attention mechanisms capable of modeling these intricate interactions. The progress in multi-scale feature extraction and the evolution of attention-based models together establish a robust theoretical foundation for the development of next-generation molecular generation frameworks.

To address these limitations, we propose MSCoD, a novel generative framework that enhances Bayesian updating using Multi-Scale Information Bottleneck (MSIB) and multi-head cooperative attention mechanism (MHCA). The MSIB module introduces semantic space compression at multiple abstraction levels, enabling efficient hierarchical feature extraction for both local atomic details and global molecular patterns. Then, the MHCA module applies asymmetric protein-to-ligand attention, specifically designed to handle the dimensionality gap between proteins and ligands and to reflect the biological adaptation of ligands to protein environments.

Experimental results demonstrate that MSCoD significantly outperforms existing methods in generating high-quality 3D molecules and improving target binding affinity across multiple evaluation metrics, validating its transformative potential for structure-based drug design applications.  All in all, the
 main contributions of this article are as follows.

\begin{itemize}
    \item \textbf{Semantic Space Compression of molecular features}: To the best of our knowledge, this is the first time that a multi-scale information bottleneck has been introduced into the field of molecular design, which better performs semantic compression at multiple levels of abstraction and achieves efficient hierarchical extraction of molecular features.
    \item \textbf{Asymmetric Cooperative Attention}: MSCoD proposes a multi-head cooperative attention mechanism to model the inherent asymmetry between proteins and ligands, enabling ligand features to adapt dynamically to the protein environment for more accurate interaction modeling.
    \item \textbf{Superior Performance}: Extensive empirical evaluations show that MSCoD outperforms state-of-the-art methods on structure-based molecular generation benchmarks. Case studies on challenging targets such as KRAS G12D (7XKJ) demonstrate its practical applicability, and the MSIB and MHCA modules transfer effectively to drug--target affinity prediction, improving predictive accuracy and robustness on standard benchmarks.
\end{itemize}

\section{\textbf{2 Materials and methods}}\label{sec2}
\subsection{\textbf{2.1 Problem formulation}}\label{subsec1}

From a generative modeling perspective, the SBDD task\citep{wang2018structure} can be defined as generating molecules that can bind to a given protein pocket with high affinity. The input is a protein binding site \( P = \{(x^{(i)}_P, v^{(i)}_P)\}_{i=1}^{N_P} \), where \( N_P \) is the number of protein atoms, each \( x^{(i)}_P \in \mathbb{R}^3\) and \( v^{(i)}_P \in \mathbb{R}^{D_P} \) represent continuous three-dimensional atomic coordinates and discrete but continuously modeled atomic features (e.g., element type, backbone or side-chain indicators), respectively. The output is a ligand molecule \( M = \{(x^{(i)}_M, v^{(i)}_M)\}_{i=1}^{N_M} \), where \( x^{(i)}_M \in \mathbb{R}^3 \) and \( v^{(i)}_M \in \mathbb{R}^{D_M} \), and \( N_M \) is the number of atoms in the molecule. For convenience, we denote \( p = [x_P, v_P] \) (where \( x_P \in \mathbb{R}^{N_P \times 3} \), \( v_P \in \mathbb{R}^{N_P \times D_P} \)) and \( m = [x_M, v_M] \) (where \( x_M \in \mathbb{R}^{N_M \times 3} \), \( v_M \in \mathbb{R}^{N_M \times D_M} \)) as concatenations of all protein atoms and concatenations of all ligand atoms, respectively.

\begin{figure*}[htbp]
    \centering
    \includegraphics[width=0.95\textwidth]{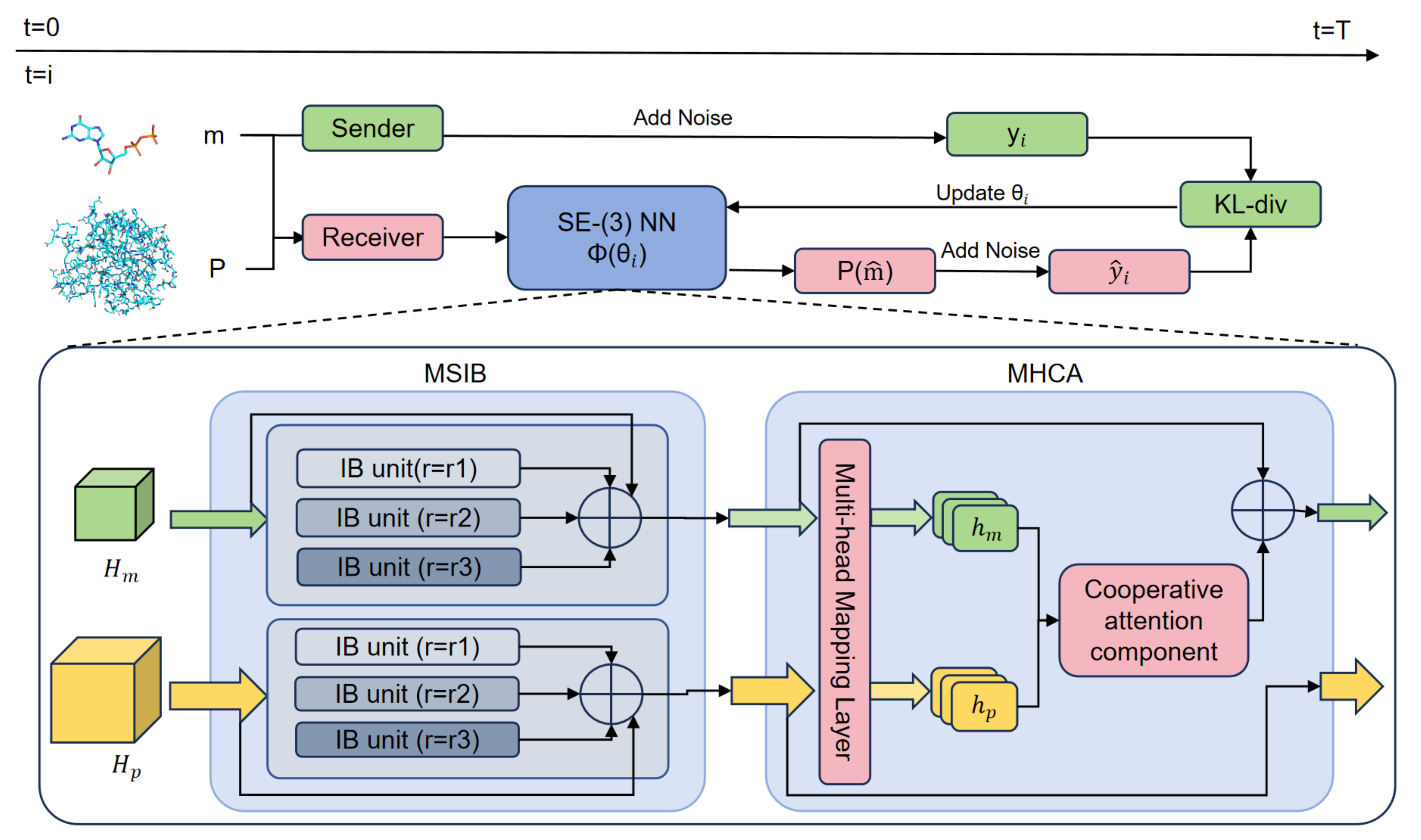}
    \caption{Overview of the MSCoD framework showing the integration of multi-scale information bottleneck and multi-head cooperative attention mechanism for enhanced molecular generation. This method combines Bayesian updating with sophisticated attention mechanisms to model protein-ligand interactions effectively.}
    \label{fig:MSCoD}
\end{figure*}

\subsection{\textbf{2.2 Overview of MSCoD}}\label{subsec2}

Figure~\ref{fig:MSCoD} presents an overview of the MSCoD framework for structure-based drug design. The process begins with the input of both protein and ligand molecules, which are processed along two parallel pathways. In the first pathway, the ligand molecule is perturbed by adding noise, generating a candidate ligand sample. In the second pathway, both the protein and ligand molecules are jointly processed to extract their features, which are then fed into a neural network to capture the probability distribution of the ligand. This distribution is used to generate another perturbed ligand sample. The compatibility between these generated ligand samples and the protein binding site is then evaluated, and the resulting information is used to update the model parameters. By iteratively repeating this process, the model progressively improves its ability to generate ligands that are highly compatible with the target protein.

Throughout this workflow, both atomic coordinates and atom types are represented in a unified feature space, enabling consistent and efficient processing. The neural network $\Phi$ generates ligand candidates based on current model parameters and protein context, while the update function $P_U$ refines the parameters according to the compatibility between generated ligands and the protein binding site:
\begin{equation}
\theta_{i-1}  \xrightarrow{\Phi} \hat{m}  \xrightarrow{p_{U}} \theta_i
\end{equation}

As illustrated in Figure~\ref{fig:MSCoD}, the framework is further enhanced by the Multi-Scale Information Bottleneck (MSIB) and the Multi-Head Cooperative Attention (MHCA) mechanism, which together enable hierarchical feature extraction and adaptive protein-ligand interaction modeling. The complete algorithmic workflow integrating MSIB and MHCA is formally described in Algorithm~\ref{algo1}.

The following subsections provide further details: Section 2.3 elaborates on the Bayesian updating principle underlying the molecular generation process; Section 2.4 introduces the MSIB module; Section 2.5 presents the MHCA mechanism; and Section 2.6 outlines the training objective. 

For clarity, our experimental evaluation first reports target-specific generation results and case studies (Section~3.5, e.g. the KRAS G12D / 7XKJ study), followed by cross-task transfer experiments for drug–target affinity prediction (Section~3.6). This ordering reflects the primary validation of the generation method first, and the subsequent transfer evaluation to affinity prediction as a separate cross-task assessment.

\begin{algorithm}[!t]
\caption{MSIB and MHCA Mechanism}\label{algo1}
\begin{algorithmic}[1]
\Require Protein features $H_{p} \in \mathbb{R}^{N_p \times d}$, ligand features $H_{m} \in \mathbb{R}^{N_l \times d}$
\Ensure Enhanced protein and ligand representations $H_{p}^{final}$, $H_{m}^{final}$

\State Initialize MSIB with bottleneck dimensions $\{r1, r2, r3\}$ and MHCA with $H$ heads
\For{$r \in \{r1, r2, r3\}$}
    \State $Z_r^{enc} \Leftarrow \text{ReLU}(W_r^{enc} H_{p} + b_r^{enc})$
    \State $Z_r^{proc} \Leftarrow \text{ReLU}(Z_r^{enc})$
    \State $Z_r^{dec} \Leftarrow \text{ReLU}(W_r^{dec} Z_r^{proc} + b_r^{dec})$
\EndFor
\State $H_{p}^{ms} \Leftarrow H_{p} + \sum_{r} Z_r^{dec}$
\State $H_{m}^{ms} \Leftarrow$ Apply steps 4-7 to $H_{m}$
\For{$h = 1$ \textbf{to} $H$}
    \State $P_h \Leftarrow W_{P_h} H_{p}^{ms}$, $L_h \Leftarrow W_{L_h} H_{m}^{ms}$
    \State $A_h \Leftarrow \text{softmax}\left(\frac{P_h \cdot L_h^T}{\sqrt{d_h}}\right)$
    \State $C_h \Leftarrow A_h^T \cdot P_h$, $G_h \Leftarrow \sigma(W_g H_{m}^{ms})$
    \State $C_h^{gated} \Leftarrow C_h \odot G_h$
    \State $U_h \Leftarrow W_u^h C_h^{gated}$
\EndFor
\State $H_{m}^{enhanced} \Leftarrow W_o \text{Concat}(U_1, U_2, ..., U_H)$
\State $H_{p}^{final} \Leftarrow \text{FFN}(\text{LayerNorm}(H_{p}^{ms}))$
\State $H_{m}^{final} \Leftarrow \text{FFN}(\text{LayerNorm}(H_{m}^{enhanced} + H_{m}^{ms}))$
\State \textbf{return} $H_{p}^{final}$, $H_{m}^{final}$
\end{algorithmic}
\end{algorithm}

\subsection{\textbf{2.3 Molecular Generation Process}}\label{subsec3}

The molecular generation process in MSCoD is fundamentally grounded in Bayesian updating theory~\citep{qu2024molcraft}. In Bayesian inference, a prior distribution is iteratively refined into a posterior by systematically incorporating new evidence. Within our framework, the model maintains a prior belief over ligand structures, which is updated to a posterior distribution as evidence from protein-ligand interactions is incorporated.

Specifically, ligand candidates are generated based on the current model parameters and protein binding site, modeling the prior distribution of compatible molecules. The compatibility between generated ligands and the protein environment provides evidence to update the model parameters, forming the posterior distribution. This iterative Bayesian updating links the candidate ligand generation and model refinement throughout the molecular generation process.

\textbf{Probabilistic Modeling:}  
To formally describe the molecular generation process, we adopt a probabilistic framework that models both atomic coordinates and atom types. 

First, the conditional likelihood of generating a molecule $y$ given latent variable $m$ is defined as:
\begin{equation}
p_S(y | m; \alpha, \alpha') = \mathcal{N}(y_x | x_M, \alpha^{-1}I) \cdot \mathcal{N}(y_v | \alpha'(Ke_{v_M} - 1), \alpha'KI)
\end{equation}
Here, $y_x$ and $y_v$ represent the atomic coordinates and atom types of the generated molecule, respectively. The likelihood is modeled as the product of two Gaussian distributions: one for the coordinates and one for the atom types. The parameters $\alpha$ and $\alpha'$ control the variance, $e_{v_M}$ encodes the atom types, $K$ is a scaling constant, and $I$ is the identity matrix, which defines the covariance structure. This formulation allows the model to capture both spatial and chemical uncertainty in the generation process.

Next, the posterior distribution over molecules is computed by integrating over the generative process:
\begin{equation}
p_R(y_i | \theta_{i-1}, p; t_i) = \mathbb{E}_{\hat{m} \sim \Phi(\theta_{i-1}, p, t_i)}[p_S(y_i | \hat{m}; \alpha_i)]
\end{equation}
where $\Phi$ denotes the neural network generator, $t_i$ is the current time step, and $\alpha_i$ is the noise schedule parameter at that step. This equation expresses the expected likelihood of a molecule $y_i$ under the distribution of candidate molecules $\hat{m}$ generated by the model, given the current parameters and protein context. It reflects how the model updates its belief about plausible ligand structures by integrating evidence from the generative process.

Finally, for the overall prediction, the model marginalizes over possible atom types to obtain the final probability:
\begin{equation}
\begin{aligned}
p_R(y | \theta, p; t) =\ & \mathcal{N}(y_x | \Phi(\theta_x, p, t), \alpha^{-1}I) \\
& \cdot \left[ \sum_k p^v_O(k|\cdot)\, p^v_S(y_v^{(d)}|k; \alpha) \right]_{d=1...N}
\end{aligned}
\end{equation}
In this formulation, $\Phi(\theta_x, p, t)$ represents the coordinate prediction part of the model, $p^v_O$ is the predicted probability for atom type $k$ and $p^v_S$ is the likelihood for atom type $k$. The summation over $k$ and $d$ (up to the number of ligand atoms $N_M$) accounts for all possible atom types and all atoms in the molecule, ensuring that the model considers the full chemical diversity in its predictions.

\textbf{Protein-Ligand Modeling:}  
In this framework, the protein binding site $p$ provides the essential structural and chemical context for ligand generation. To effectively model protein-ligand interactions, our approach leverages the multi-scale information bottleneck modules (Section 2.4) and the multi-head cooperative attention mechanism (Section 2.5). These modules jointly extract hierarchical features and capture asymmetric adaptation processes, ensuring that ligand candidates are not only structurally plausible but also compatible with the protein environment. As a result, the posterior distribution favors ligands with high binding affinity and desirable pharmaceutical properties, enabling efficient exploration and adaptive optimization within chemical space.

\textbf{SE(3)-Equivariant Neural Network}
Moreover, To generate 3D molecules from protein binding sites, the model must predict both atomic coordinates and atom types while maintaining SE(3)-equivariance, ensuring invariance to translation and rotation~\citep{schneuing2024structure, guan3d2023, pmlr-v202-guan23a}. This property guarantees physically valid molecular structures. Formally:

\textbf{Proposition 1.} For any SE(3) transformation $T_g$, the likelihood is invariant: $p_\phi(T_g(m|p)) = p_\phi(m|p)$, provided the protein center of mass is shifted to zero and $\Phi(\theta, p, t)$ is SE(3)-equivariant.

We implement SE(3)-equivariant neural networks on $k$-nearest neighbor graphs of protein-ligand complexes to jointly learn atomic coordinates and types. At each layer $l$, hidden embeddings $h$ and coordinates $x$ are updated as:
\begin{align}
h^{l+1}_i &= h^l_i + \sum_{j \in \mathcal{N}_G(i)} \phi_h\left(d^l_{ij}, h^l_i, h^l_j, e_{ij}, t\right) \\
\Delta x_i &= \sum_{j \in \mathcal{N}_G(i)} (x^l_j - x^l_i)\, \phi_x\left(d^l_{ij}, h^{l+1}_i, h^{l+1}_j, e_{ij}, t\right) \\
x^{l+1}_i &= x^l_i + \Delta x_i \cdot \mathds{1}_{mol}
\end{align}
where $\mathcal{N}_G(i)$ denotes the neighborhood of atom $i$ in $G$, $d^l_{ij} = \|x^l_i - x^l_j\|$ is the Euclidean distance between atoms $i$ and $j$, and $e_{ij}$ indicates whether the edge is between protein atoms, ligand atoms, or protein-ligand atom pairs. The indicator $\mathds{1}_{mol}$ ensures that only ligand atom coordinates are updated. The functions $\phi_h$ and $\phi_x$ are attention blocks, where $\phi_h$ takes $h^l_i$ as query and $[h^l_i, h^l_j, e_{ij}]$ as keys and values.

This design allows spatial information to be encoded in hidden features $h$, enabling downstream modules (MSIB and MHCA) to update $h$ without explicit coordinate operations. Specifically, In the MSCoD framework, we strategically focus on updating atomic feature embeddings rather than explicit atomic coordinates during intermediate layers. This approach allows downstream modules (MSIB and MHCA) to further optimize these features without direct coordinate manipulation. Such a design not only preserves SE(3)-equivariance, but also improves computational efficiency and supports effective modeling of complex protein-ligand interactions in the latent feature space, fully aligning with the overall generative modeling strategy.

\begin{figure*}[htbp]
    \centering
    \includegraphics[width=0.95\textwidth]{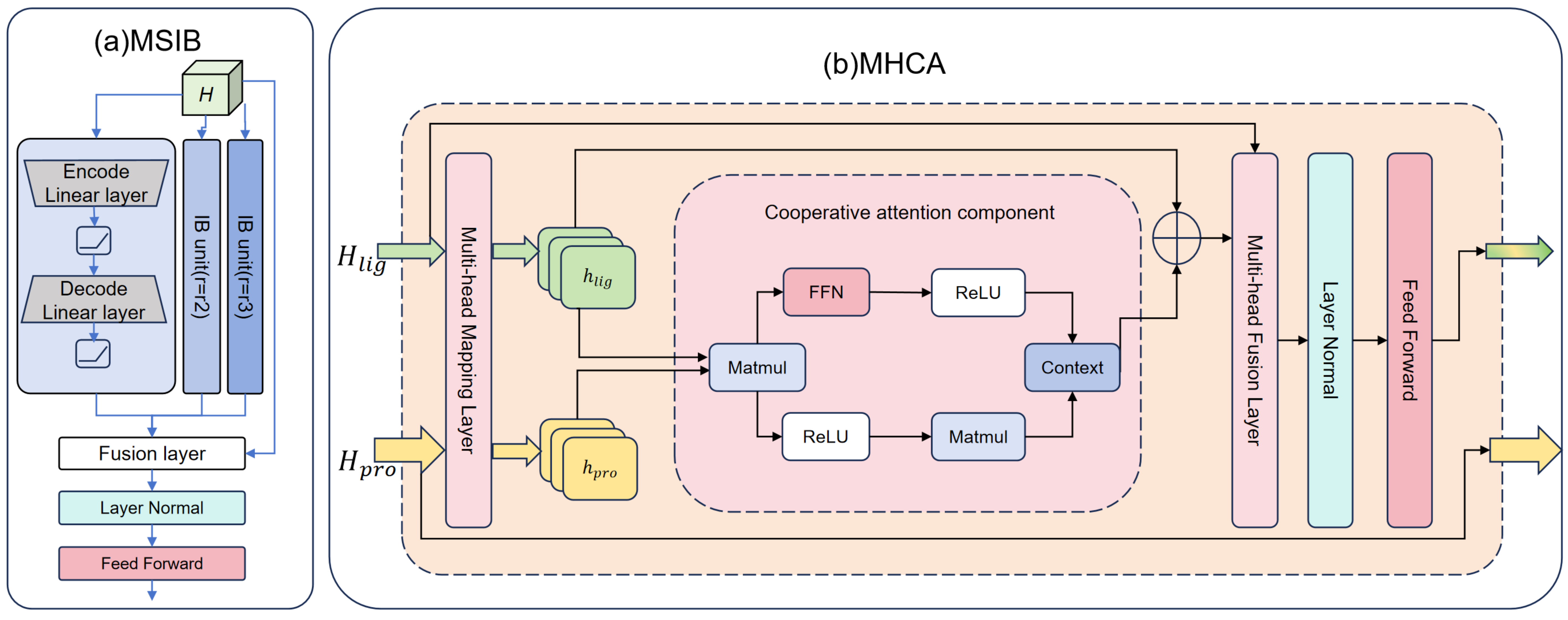}
    \caption{Detailed architecture of MSCoD modules. 
    (a) Multi-Scale Information Bottleneck (MSIB) modules: illustrates the hierarchical semantic compression and feature fusion process for protein and ligand representations. 
    (b) Multi-Head Cooperative Attention (MHCA) mechanism: shows the asymmetric attention flow from protein to ligand features, the multi-head projection strategy, dynamic gating mechanisms, and the residual processing pipeline. Each attention head specializes in capturing different aspects of protein-ligand interactions, with the final integration producing enhanced ligand representations that better reflect binding site compatibility and molecular optimization requirements.}
    \label{fig:MSIB and MHCA Mechanism}
\end{figure*}

\subsection{\textbf{2.4 Multi-scale information-bottleneck module}}\label{subsec4}

Protein-ligand drug design requires modeling molecular interactions across multiple spatial scales. Proteins and ligands exhibit hierarchical structures, where local atomic contacts, intermediate functional groups, and global conformations all contribute to binding affinity and selectivity. For example, protein active sites are defined by local geometric and chemical properties, while ligand binding depends on both global shape complementarity and overall molecular architecture. Traditional single-scale feature extraction methods~\citep{kawaguchi2023does, hu2024survey} often fail to capture this complexity, resulting in incomplete representations and suboptimal generation.

Inspired by feature pyramid networks and UNet architectures in computer vision~\citep{cao2022swin}, we propose Multi-Scale Information Bottleneck (MSIB) modules to systematically extract features at multiple abstraction levels. Each bottleneck layer adopts a U-shaped encoder-decoder structure with downsampling, nonlinear transformation, and upsampling, tailored for molecular representation learning. This design enables the model to capture both fine-grained atomic details and global molecular patterns essential for accurate binding affinity prediction. The detailed architecture is shown in Figure~\ref{fig:MSIB and MHCA Mechanism}a.

Additionally, our approach shifts from traditional physical space compression to semantic space compression. Unlike conventional methods that reduce dimensionality by sampling or clustering atoms—potentially losing critical binding information—MSIB operates in the latent feature space, compressing feature dimensions while preserving molecular connectivity. This transition improves both computational efficiency and feature quality, as semantic compression maintains the complete molecular structure and extracts hierarchical representations at different abstraction levels.

Given input hidden embeddings $H \in \mathbb{R}^{N \times d}$, where $N$ is the number of atoms and $d$ the feature dimension, we define three parallel pathways with different compression ratios. For each pathway $r$, the processing follows a unified encoder-decoder structure:

\begin{align}
Z_r^{enc} &= \text{ReLU}(W_r^{enc} H + b_r^{enc}) \\
Z_r^{proc} &= \text{ReLU}(Z_r^{enc}) \\
Z_r^{dec} &= \text{ReLU}(W_r^{dec} Z_r^{proc} + b_r^{dec})
\end{align}
where $r \in \{r1, r2, r3\}$ denotes different feature dimension compression ratios, forming a feature pyramid. The parameters $W_r^{enc}, W_r^{dec}$ and $b_r^{enc}, b_r^{dec}$ are learnable weights and biases specific to each compression ratio $r$.

Each bottleneck dimension targets a specific biological scale:
\begin{itemize}
\item \textbf{High compression ratio($r=r1$)}: Captures global semantic patterns for overall binding compatibility, forcing the model to focus on fundamental molecular descriptors.
\item \textbf{Medium compression ratio($r=r2$)}: Balances local atomic details and broader structural patterns, preserving intermediate functional group semantics critical for specificity and selectivity.
\item \textbf{Low compression ratio($r=r3$)}: Retains fine-grained information, such as local atomic interactions, hydrogen bonds, and electrostatic complementarity, essential for accurate binding strength prediction.
\end{itemize}

Unlike UNet's concatenation-based skip connections, which increase dimensionality, we use additive feature fusion for efficient integration of multi-scale representations:
\begin{equation}
H_{fusion} = H_{original} + \sum_{r \in \{r1,r2,r3\}} Z_r^{dec}
\end{equation}

The fused features are then normalized and processed by a two-layer feed-forward network with ReLU activation:
\begin{align}
H_{norm} &= \text{LayerNorm}(H_{fusion}) \\
H_{output} &= \text{FFN}(\text{ReLU}(\text{FFN}(H_{norm})))
\end{align}

This semantic multi-scale approach enables the model to capture both the atomic details and global patterns essential for binding affinity prediction and high-quality molecular generation, while maintaining computational efficiency through an intelligent feature space.

\subsection{\textbf{2.5 Multi-Head Cooperative Attention Mechanism}}\label{subsec5}

Traditional attention mechanisms~\citep{vaswani2017attention,huang2022coadti} in molecular modeling typically focus on intra-molecular relationships and treat protein-ligand interactions symmetrically, overlooking the inherent asymmetry in binding processes. In structure-based drug design, however, a fundamental biological asymmetry exists: protein binding sites provide relatively stable structural templates, while ligands must adapt and optimize their conformations for optimal binding.

To address this, we propose a Multi-Head Cooperative Attention (MHCA) mechanism that enhances ligand representations using protein context, reflecting the natural adaptation of ligands to pre-existing protein environments. The detailed architecture is shown in Figure~\ref{fig:MSIB and MHCA Mechanism}b. Unlike traditional attention, our approach is engineered to capture complex physicochemical interactions between proteins and ligands, with each attention head specializing in a distinct interaction modality—such as electrostatic complementarity, hydrophobic interactions, hydrogen bonding, or shape matching. Unlike conventional cross-attention, which assumes symmetric feature dimensions, our mechanism addresses the challenge of large atomic count disparities between proteins (200-2000 atoms) and ligands (20-100 atoms). We achieve this through specialized projection strategies and dimension-aware processing, allowing the framework to adapt dynamically to asymmetric feature spaces. The unidirectional attention guidance aligns with the biological reality: ligands adapt to protein binding sites, not vice versa. This design enhances ligand optimization while maintaining protein structural stability, reflecting the true asymmetry of drug design.

Given protein features $H_{protein} \in \mathbb{R}^{N_p \times d}$ and ligand features $H_{ligand} \in \mathbb{R}^{N_l \times d}$, we project these into multiple attention subspaces to systematically capture diverse interaction patterns. For each attention head $h \in \{1, 2, ..., H\}$, we perform independent linear transformations without bias terms:

\begin{align}
P_h &= W_{P_h} H_{protein}, \quad P_h \in \mathbb{R}^{N_p \times d_h} \\
L_h &= W_{L_h} H_{ligand}, \quad L_h \in \mathbb{R}^{N_l \times d_h}
\end{align}
where $W_{P_h}, W_{L_h} \in \mathbb{R}^{d_h \times d}$ are learnable projection matrices and $d_h = d / H$ is the dimension per head. The absence of bias terms ensures projections focus on pure feature relationships, reducing overfitting and improving generalization.

The asymmetric attention mechanism determines how each ligand atom attends to protein atoms to optimize binding. For each head, we compute the protein context influencing ligand optimization, effectively creating a "protein fingerprint" for each ligand atom:

\begin{align}
\text{Attention}_{h} &= \text{softmax}\left(\frac{P_h \cdot L_h^T}{\sqrt{d_h}}\right) \in \mathbb{R}^{N_p \times N_l} \\
C_h &= \text{Attention}_{h}^T \cdot P_h \in \mathbb{R}^{N_l \times d_h}
\end{align}

The scaling factor $\sqrt{d_h}$ stabilizes gradients during training. The resulting attention matrix captures the interaction potential between each protein-ligand atom pair, where $\text{Attention}_{h}[i,j]$ reflects the relevance of protein atom $i$ to ligand atom $j$.

To enable selective information incorporation and prevent information overflow, we implement a gating mechanism that allows ligand atoms to incorporate the relevant protein context selectively:

\begin{align}
G_h &= \sigma(W_g H_{ligand}) \in \mathbb{R}^{N_l \times d_h} \\
C_h^{gated} &= C_h \odot G_h \\
U_h &= W_u^h C_h^{gated} \in \mathbb{R}^{N_l \times d}
\end{align}
where $\sigma$ is the Sigmoid activation, $W_g \in \mathbb{R}^{d_h \times d}$ is a learnable gating matrix, $\odot$ denotes element-wise multiplication, and $W_u^h \in \mathbb{R}^{d \times d_h}$ is a head-specific transformation matrix. This gating mechanism filters out irrelevant protein information, allowing different ligand atoms to respond variably to protein context, while each attention head generates an enhanced ligand representation through its dedicated update layer.

Subsequently, outputs from all heads are concatenated and fused through a learned integration layer, followed by residual processing for stability:

\begin{align}
H_{ligand}^{attended} &= W_o \text{Concat}(U_1, U_2, ..., U_H) \\
H_{ligand}^{enhanced} &= \text{LayerNorm}(H_{ligand}^{attended} + H_{ligand}) \\
H_{ligand}^{final} &= \text{FFN}(\text{ReLU}(\text{FFN}(H_{ligand}^{enhanced})))
\end{align}
where $W_o \in \mathbb{R}^{d \times (H \times d)}$ integrates multi-head representations, combining diverse interaction patterns into a coherent enhanced representation. The residual connection preserves original ligand information while incorporating protein-aware enhancements, followed by normalization and feed-forward processing for feature refinement.

\subsection{\textbf{2.6 Training Objective}}\label{subsec6}

To train MSCoD, we adopt the objective function from the Bayesian Flow Network (BFN) framework, following the approach of MolCRAFT~\citep{qu2024molcraft}. This involves optimizing parameters by minimizing the KL-divergence between noisy sample distributions. While BFN accommodates both discrete-time and continuous-time regimes, we adopt the $n$-step discrete formulation for computational efficiency. Since atomic coordinates and noise follow Gaussian distributions, closed-form loss derivations are feasible.

The coordinate loss function is formulated as:
\begin{align}
L_n^x =\ & D_{\mathrm{KL}}\left(\mathcal{N}(x, \alpha_i^{-1} I)\ \|\ \mathcal{N}(\hat{x}(\theta_{i-1}, p, t), \alpha_i^{-1} I)\right) \notag \\
&= \frac{\alpha_i}{2} \| x - \hat{x}(\theta_{i-1}, p, t) \|^2
\end{align}

For atom type prediction, we compute the corresponding KL-divergence:
\begin{align}
L_n^v =\ & \ln \mathcal{N}\left(y^v \mid \alpha_i (K_{ev} - 1), \alpha_i K I\right) \notag \\
& - \sum_{d=1}^{N_M} \ln \left( \sum_{k=1}^{K} p_O(k|\theta; t)\, \mathcal{N}\left(y_v^{(d)} \mid \alpha_i (K_{ek} - 1), \alpha_i K I\right) \right)
\end{align}

Here, $x$ represents the ground truth coordinates, $\hat{x}(\theta_{i-1}, p, t)$ denotes predicted coordinates, $\alpha_i$ controls noise at step $i$, $y_v$ indicates atom types, $p_O(k|\theta; t)$ provides type probabilities, $K_{ek}$ encodes one-hot representations, with $N_M$ and $K$ representing atom and type counts respectively.

By exploiting analytical KL-divergence formulations, this framework ensures efficient and stable MSCoD training. The complete objective combines both loss components:
\begin{equation}
L_{\text{total}} = L_n^x + L_n^v
\end{equation}

\section{\textbf{3 Experimental Settings}}\label{sec3}
\subsection{\textbf{3.1 Datasets and Parameter Setting}}
For molecular generation evaluation, following previous work\citep{luo2021AR,guan3d2023,peng2022pocket2mol}, we train and evaluate MSCoD on the CrossDocked2020 dataset\citep{francoeur2020three}. This dataset is widely used for benchmarking SBDD methods. We follow the same data preparation and splitting method as \citep{luo2021AR}. 22.5 million docked binding complexes are refined to high-quality docking poses with protein diversity to ensure comprehensive evaluation. This yields 100,000 protein-ligand pairs for training and 100 proteins for testing with diverse binding sites. All sampling and evaluation procedures follow the method of \citep{guan3d2023} to ensure fair comparison with existing methods.

For the MSCoD architecture, we implement MSIB with feature dimension compression ratios of $\{0.125, 0.25, 0.5\}$ and configure MHCA with 8 attention heads. The hidden dimension is set to 128 with attention dimension calculated as $d_h = d/H = 128/8 = 16$ per head. We employ ReLU activation functions with Layer Normalization throughout the network. The model incorporates residual connections and feedforward neural networks for feature enhancement and stabilization. For training, we adopt the Adam optimizer with learning rate 0.005, batch size of 4, and train the model for 25 epochs. The training converges within the specified epochs on our workstation equipped with an Intel Core i7-14700KF processor (20 cores, 28 threads) and NVIDIA GeForce RTX 4060 Ti with 16GB VRAM. Training requires approximately 5 days for completion. All experiments are conducted with 64GB of system memory to provide comprehensive computational support. The model parameters are updated through standard backpropagation with gradient clipping to ensure training stability.

\begin{table*}[htbp]
\centering
\setlength{\tabcolsep}{2pt}
\begin{tabular*}{\textwidth}{@{\extracolsep{\fill}}l|cc|cc|cc|ccc|c|c|c|c|c|c|c@{\extracolsep{\fill}}}
\toprule
& \multicolumn{6}{c|}{Binding Affinity} & \multicolumn{5}{c|}{Conformation Stability} & \multicolumn{3}{c|}{Drug-like Properties} & \multicolumn{2}{c}{Overall}\\
\midrule
\multirow{2}{*}{Methods} & \multicolumn{2}{c|}{Vina Score(↓)} & \multicolumn{2}{c|}{Vina Min(↓)} & \multicolumn{2}{c|}{Vina Dock(↓)} & \multicolumn{3}{c|}{SE(↓)} &{Clash(↓)} &{RMSD(↑)} &{QED(↑)} & {SA(↑)} & {Div(↑)} & {BF(↑)} & {SR(↑)}\\
 & Avg. & Med. & Avg. & Med. & Avg. & Med. & 25\% &50\% &75\% & Avg. & $<$2\AA & Avg. & Avg. & Avg. & (\%) & (\%)\\
\midrule
Reference & -6.36 & -6.46 & -6.71 & -6.49 & -7.45 & -7.26 & 34 & 107 & 196 & 5.51 & 34.0 & 0.48 & 0.73 & - & 29.0 & 25.0\\
\midrule
LiGAN & - & - & - & - & -6.33 & -6.20 & - & - & - & - & - & 0.39 & 0.59 & 0.66 & - & -\\
GraphBP & - & - & - & - & -4.80 & -4.70 & - & - & - & - & - & 0.43 & 0.49 & \textbf{0.79} & - & -\\
AR & -5.75 & -5.64 & -6.18 & -5.88 & -6.75 & -6.62 & 259 & 595 & 2286 & \textbf{4.49} & 31.1 & \underline{0.51} & 0.63 & 0.70 & 17.3 & 7.1\\
Pocket2Mol & -5.14 & -4.70 & -6.42 & -5.82 & -7.15 & -6.79 & 102 & 189 & 374 & \underline{6.24} & 30.8 & \textbf{0.56} & \textbf{0.74} & 0.69 & 24.6 & 24.4\\
DiffSBDD & -1.94 & -4.24 & -5.85 & -5.94 & -7.00 & -6.90 & - & - & - & - & - & 0.48 & 0.58 & 0.73 & - & -\\
TargetDiff & -5.47 & -6.30 & -6.64 & -6.83 & -7.80 & -7.91 & 369 & 1243 & 13871 & 10.84 & 29.4 & 0.48 & 0.58 & 0.72 & 14.4 & 10.5\\
Decomp-R & -5.19 & -5.27 & -6.03 & -6.00 & -7.03 & -7.16 & 115 & 421 & 1424 & 8.16 & 22.7 & 0.45 & 0.61 & {0.68} & 14.6 & 14.9\\
Decomp-O & -5.67 & -6.04 & -7.04 & -6.91 & \underline{-8.39} & \textbf{-8.43} & 379 & 983 & 4133 & 14.63 & 23.9 & 0.45 & 0.61 & {0.68} & 11.1 & 24.5\\
BINDDM & -5.92 & -6.81 & \underline{-7.29} & \underline{-7.34} & \textbf{-8.41} & \underline{-8.37} & 520 & 5790 & 659900 & 8.50 & 38.6 & \underbar{0.51} & 0.58 & \underline{0.75} & 8.3 & 13.7\\
MolCRAFT & \underline{-6.59} & \underline{-7.04} & -7.27 & -7.26 & -7.92 & -8.01 & \underline{83} & \underline{195} & \underline{510} & 7.09 & \underline{41.8} & 0.50 & \underline{0.69} & 0.72 & \underline{29.0} & \underline{26.8}\\
MSCoD & \textbf{-7.11} & \textbf{-7.20} & \textbf{-7.68} & \textbf{-7.48} & -8.18 & -8.19 & \textbf{32} & \textbf{96} & \textbf{252} & 8.37 & \textbf{47.7} & \textbf{0.56} & \underline{0.69} & 0.72 & \textbf{43.3} & \textbf{29.1}\\
\botrule
\end{tabular*}
\caption{Summary of different properties of reference molecules and molecules generated by our model and other baselines. (↑) / (↓) denotes a larger / smaller number is better. Top 2 results are highlighted with bold text and underlined text, respectively.}
\label{table:summary}
\end{table*}

\begin{table}[t]
\centering
\begin{tabular*}{\columnwidth}{@{\extracolsep{\fill}}l|cc|cc|c|c@{\extracolsep{\fill}}}
\toprule
\multirow{2}{*}{Methods} & \multicolumn{2}{c|}{Vina Score(↓)} & \multicolumn{2}{c|}{Vina Min(↓)} & {QED(↑)} & {SA(↑)}\\
 & Avg. & Med. & Avg. & Med. & Avg. & Avg.\\
\midrule
Baseline & -6.59 & -7.04 & \underline{-7.27} & -7.26 & 0.50 & \textbf{0.69} \\
MSIB & \underline{-6.68} & -7.12 & -7.23 & \underline{-7.37} & \underline{0.54} & \underline{0.68} \\
MHCA & -6.39 & \underline{-7.18} & -6.97 & -7.24 & 0.52 & \textbf{0.69} \\
MSCoD & \textbf{-7.11} & \textbf{-7.20} & \textbf{-7.68} & \textbf{-7.48} & \textbf{0.56} & \textbf{0.69} \\
\botrule
\end{tabular*}
\caption{Effect of the MSCoD Interaction Mechanism on Target-specific Molecule Generation}
\label{table:MSCoD-Interaction-Mechanism}
\end{table}

\begin{table}[t]
\centering
\begin{tabular*}{\columnwidth}{@{\extracolsep{\fill}}c|cc|cc|c|c@{\extracolsep{\fill}}}
\toprule
\multirow{2}{*}{Head} & \multicolumn{2}{c|}{Vina Score(↓)} & \multicolumn{2}{c|}{Vina Min(↓)} & {QED(↑)} & {SA(↑)}\\
 & Avg. & Med. & Avg. & Med. & Avg. & Avg.\\
\midrule
4 & \underline{-6.67} & \underline{-7.10} & \underline{-7.31} & -7.29 & 0.53 & 0.62 \\
8 & \textbf{-7.11} & \textbf{-7.20} & \textbf{-7.68} & \textbf{-7.48} & \textbf{0.56} & \textbf{0.69} \\
16 & -6.70 & -6.82 & -7.22 & \underline{-7.33} & \underline{0.54} & \underline{0.67} \\
\botrule
\end{tabular*}
\caption{Influence of the number of attention heads in MHCA on Target-specific Molecule Generation}
\label{table:attention-heads}
\end{table}

\subsection{\textbf{3.2 Evaluation Metrics}}
We comprehensively evaluate generated molecules from three aspects: target binding affinity, molecular properties, and molecular structure quality. To estimate target binding affinity accurately, following previous work\citep{luo2021AR,guan3d2023,peng2022pocket2mol}, we employ AutoDock Vina\citep{eberhardt2021autodock,seeliger2010ligand} to calculate and report binding-related metrics for comprehensive assessment. Vina Score directly estimates binding affinity based on generated 3D molecules without additional processing. Vina Min performs local structure minimization before estimation to account for conformational flexibility. Vina Dock involves additional re-docking processes and reflects the best possible binding affinity achievable.

For molecular property evaluation, we use QED, SA, and Diversity as metrics, following\citep{luo2021AR,guan3d2023,peng2022pocket2mol} for consistent comparison. QED is a quantitative estimate of drug-likeness combining multiple desirable molecular properties including molecular weight, lipophilicity, and toxicity indicators. SA (Synthetic Accessibility) is a measure of the difficulty of ligand synthesis in practical laboratory settings. Diversity is calculated as the average pairwise dissimilarity among all generated ligands to assess chemical space coverage.

For molecular structure evaluation, we use RMSD as a metric for geometric accuracy assessment. RMSD evaluates the similarity between molecules by calculating the deviation between atom positions in a molecular model and corresponding atoms in the reference structure. In molecular docking or molecular simulation applications, RMSD values less than 2\AA are generally considered ideal for practical applications\citep{sargsyan2017molecular, alhossary2015fast, mcnutt2021gnina}. For binding complex, we adopt Steric Clashes (Clash) to detect possible clashes in protein-ligand complex, following \citep{harris2023posecheck}. To evaluate the overall quality of the generated molecules, we calculate Binding Feasibility as the ratio of molecules that simultaneously exhibit reasonable affinity (Vina Score $<$ -2.49 kcal/mol) and a stable conformation (Strain Energy $<$ 836 kcal/mol, RMSD $<$ 2\AA). The threshold values are set to the 95th percentile of the reference molecules. We also report Success Rate (Vina Dock $<$ -8.18, QED $>$ 0.25, SA $>$ 0.59) following \citep{long2022zero} and \citep{guan3d2023}.

\begin{figure*}[htbp]
    \centering
    \includegraphics[width=0.95\textwidth]{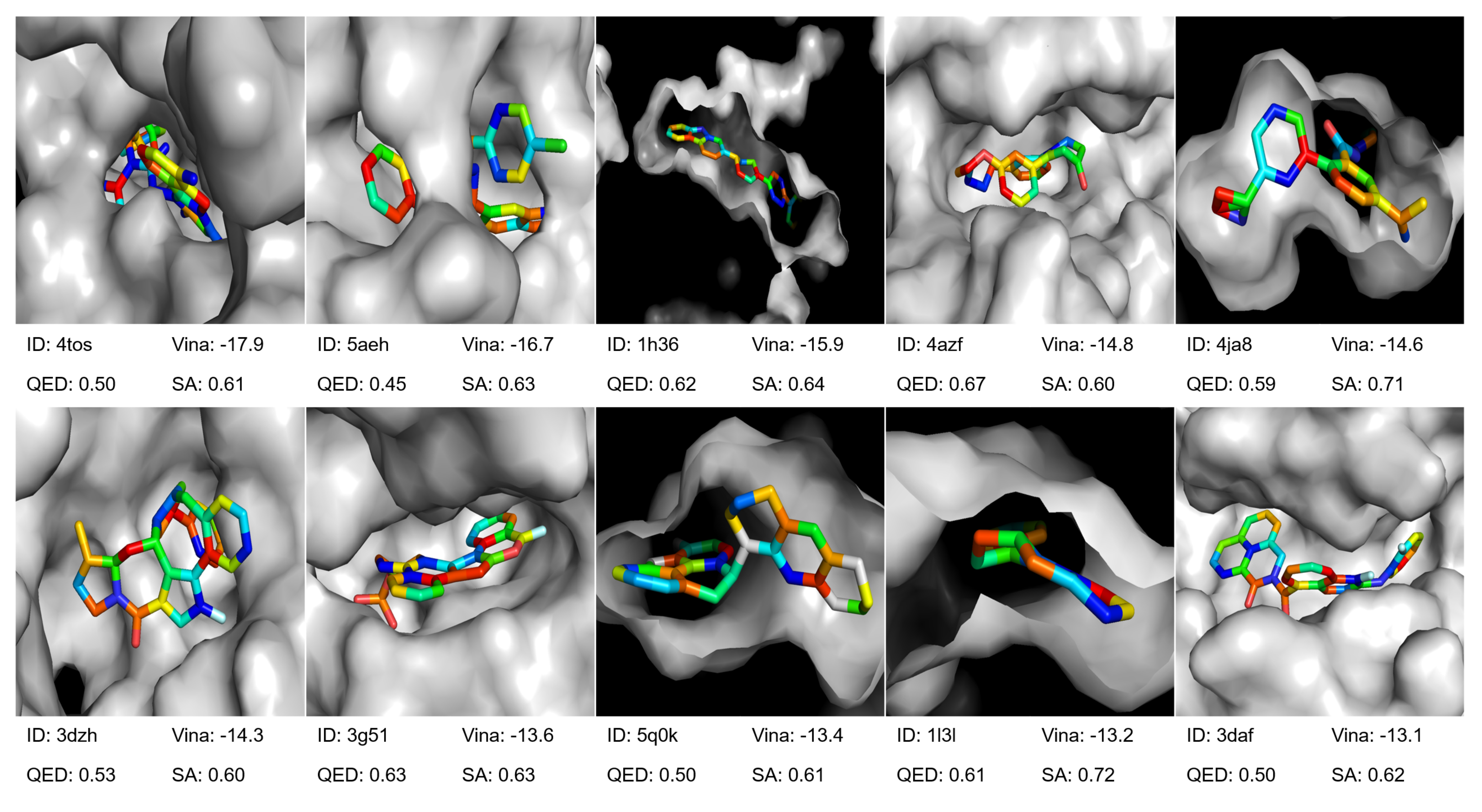}
    \caption{Ten representative ligand molecules generated by MSCoD with excellent QED and SA properties and superior Vina docking scores.}
    \label{fig:results}
\end{figure*}

\subsection{\textbf{3.3 Method Comparison}}
We compare our model with 10 recent representative SBDD methods across different algorithmic paradigms. LiGAN\citep{ragoza2022ligan} is a conditional variational autoencoder (VAE) model trained on atomic density grid representations of protein-ligand structures for 3D molecular generation. AR\citep{luo2021AR}, GraphBP\citep{liu2022Graphbp}, and Pocket2Mol\citep{peng2022pocket2mol} are autoregressive methods that generate 3D molecular atoms based on protein pockets and previously generated atoms in sequential manner. TargetDiff\citep{guan3d2023} , DecompDiff-R\citep{pmlr-v202-guan23a}, DecompDiff-O\citep{pmlr-v202-guan23a}, and BINDDM\citep{huang2024binding} are recent state-of-the-art diffusion methods that generate atomic coordinates and atom types in a non-autoregressive manner with sophisticated noise scheduling. MolCRAFT\citep{qu2024molcraft} is a recent Bayesian updating-based molecular generation method that employs a fully continuous parameter space operation approach, significantly reducing input variance through smooth transformations. All baselines are considered in Table~\ref{table:summary} for a comprehensive comparison. Addtionally, Figure~\ref{fig:results} presents ten representative examples of high-quality molecules generated by our MSCoD method, carefully selected to demonstrate the practical effectiveness of our approach in producing drug-like compounds with excellent binding properties and favorable pharmacological characteristics.

\subsubsection{Target Binding Affinity}
We systematically evaluated MSCoD's binding affinity performance through comprehensive comparison with established benchmark methods across multiple evaluation metrics. As demonstrated in Table~\ref{table:summary}, MSCoD exhibits superior performance in both binding affinity prediction and drug design properties. 

Specifically, MSCoD achieved a Vina Score average of -7.11 kcal/mol and median of -7.20 kcal/mol, coupled with a Vina Min average of -7.68 kcal/mol and median of -7.48 kcal/mol. These results establish MSCoD as the top-performing method among all evaluated approaches, substantially outperforming established methods such as MolCRAFT and BINDDM. This superior performance demonstrates that the generated molecules possess enhanced binding capabilities within target protein binding sites, as reflected by improved affinity predictions. 

Moreover, regarding Vina Dock scores, MSCoD achieved an average of -8.18 kcal/mol and median of -8.19 kcal/mol, indicating consistent docking stability and realistic binding conformations. In terms of drug design properties, MSCoD maintained exceptional competitiveness across multiple evaluation criteria. The achieved QED score of 0.56 ranks among the highest performers alongside the autoregressive method Pocket2Mol, indicating superior drug-likeness of the generated molecules. Additionally, MSCoD maintained competitive synthetic accessibility with an SA score of 0.69, demonstrating strong synthetic feasibility for practical pharmaceutical applications.

\begin{figure*}[htbp]
  \centering
  \includegraphics[width=0.96\textwidth]{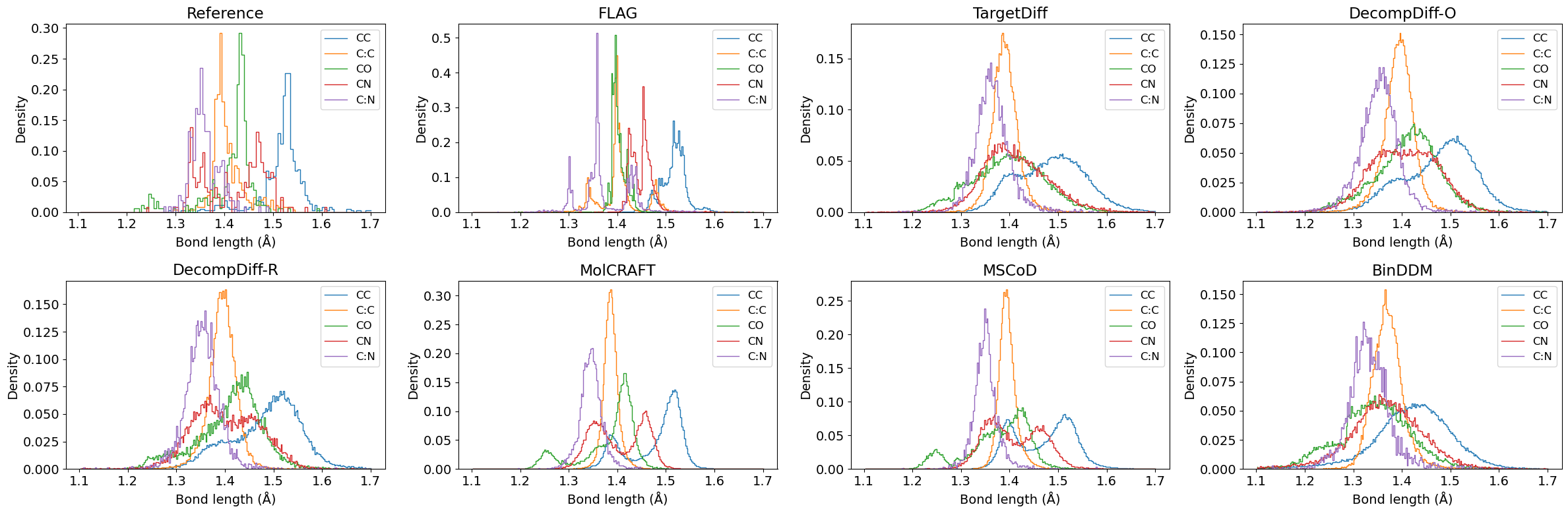}
  \caption{Bond length distribution of reference and generated molecules by autoregressive models (upper row) and nonautoregressive models (lower row) for top-5 frequent bond types}
  \label{fig:bond-length}
\end{figure*}

\begin{figure*}[htbp]
  \centering
  \begin{minipage}{0.32\textwidth}
    \centering
    \includegraphics[width=\linewidth]{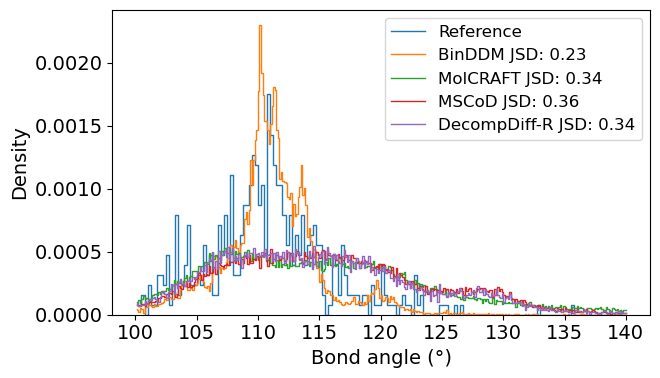}
    (C-C-C Bond Angle)
  \end{minipage}\hspace{0.01\textwidth}
  \begin{minipage}{0.32\textwidth}
    \centering
    \includegraphics[width=\linewidth]{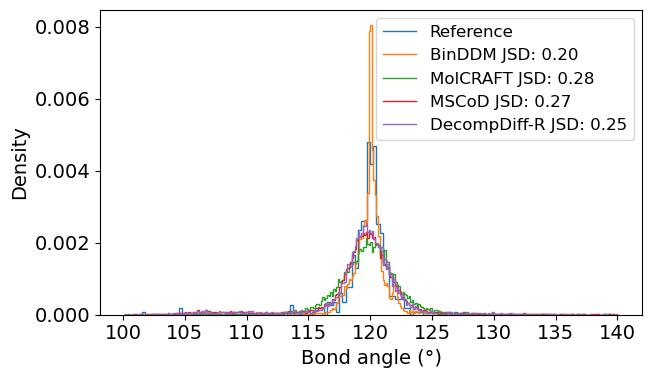}
    (C:C:N Bond Angle)
  \end{minipage}\hspace{0.01\textwidth}
  \begin{minipage}{0.32\textwidth}
    \centering
    \includegraphics[width=\linewidth]{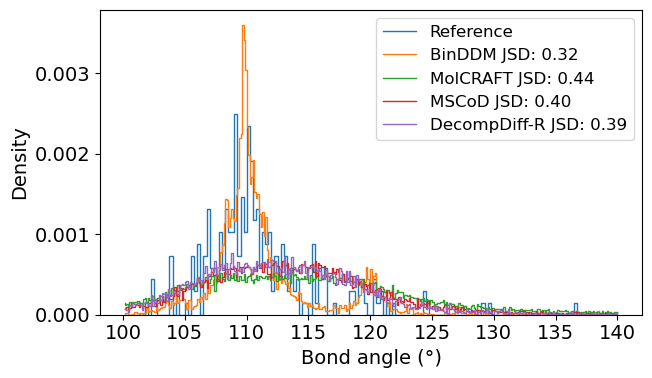}
    (C-C-O Bond Angle)
  \end{minipage}
  \caption{Bond angle distributions of generated molecules compared with reference molecules.}
  \label{fig:bond-angle}
\end{figure*}

\begin{figure*}[htbp]
  \centering
  \begin{minipage}{0.32\textwidth}
    \centering
    \includegraphics[width=\linewidth]{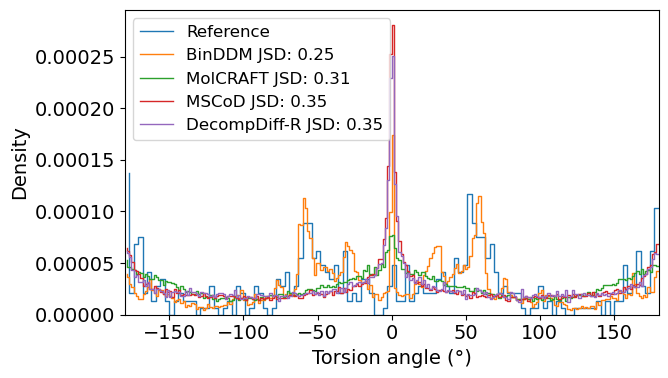}
    (C-C-C-C Torsion Angle)
  \end{minipage}\hspace{0.02\textwidth}
  \begin{minipage}{0.32\textwidth}
    \centering
    \includegraphics[width=\linewidth]{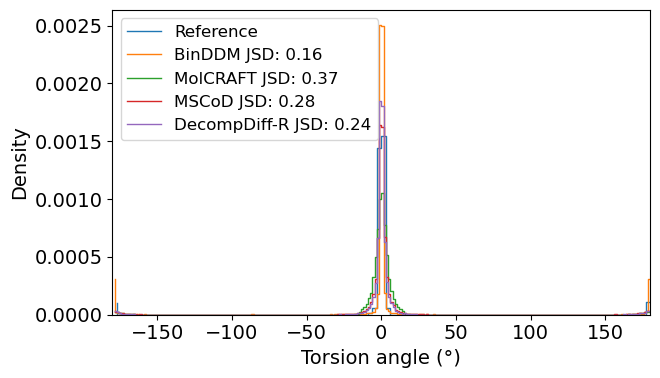}
    (C:C:C:C Torsion Angle)
  \end{minipage}
  \caption{Torsion angle distributions of generated molecules compared with reference molecules.}
  \label{fig:torsion-angle}
\end{figure*}

\subsubsection{Molecular Properties and Molecular Structures}
Regarding drug-likeness properties, MSCoD achieved a QED score of 0.56, matching the best performance with Pocket2Mol and significantly outperforming diffusion-based methods such as BINDDM (0.51) and TargetDiff (0.48). The SA score of 0.69 remained competitive with leading methods, indicating favorable synthetic feasibility. While GraphBP achieved the highest diversity score (0.79), MSCoD maintained reasonable chemical diversity (0.72) comparable to other top-performing methods including TargetDiff and MolCRAFT.

In molecular structure evaluation, MSCoD demonstrated exceptional geometric accuracy, with 47.7\% of generated structures exhibiting RMSD values below 2Å, representing the highest performance among all evaluated methods. This significantly outperformed the previous best method MolCRAFT (41.8\%) and demonstrated substantial improvement over other approaches.

Regarding conformation stability, MSCoD showed outstanding performance in ligand steric energy with the optimal percentile values (32 for 25\%, 96 for 50\%, 252 for 75\%), indicating the generation of molecules with favorable internal geometries. Although MSCoD's clash score (8.37) was moderate compared to the best performer AR (4.49), it maintained competitive performance with current state-of-the-art methods, reflecting a balanced approach between structural plausibility and binding compatibility.

The comprehensive evaluation demonstrates MSCoD's exceptional performance across multiple critical dimensions. Our method achieved the highest Binding Feasibility (43.3\%) and Success Rate (29.1\%), substantially outperforming the second-best method MolCRAFT by 49.3\% and 8.6\%, respectively. This superior performance in overall success metrics, combined with state-of-the-art binding affinity and geometric accuracy, positions MSCoD as a significant advancement in computational drug design methodology.

Additionally, we also analyzed the distributions of bond lengths, bond angles, and torsion angles of the generated ligands to assess structural plausibility. Figure~\ref{fig:bond-length} compares bond-length distributions for the five most frequent bond types; the distributions of MSCoD-generated molecules closely match those of the reference set across all bond classes, indicating accurate control of interatomic distances. Figures~\ref{fig:bond-angle} and \ref{fig:torsion-angle} present bond-angle and torsion-angle distributions, respectively; the generated molecules show strong agreement with reference distributions for these geometric descriptors, further validating the model's ability to reproduce realistic 3D geometry. Together, these results demonstrate that MSCoD not only improves binding affinity and drug-like properties but also generates chemically and geometrically plausible molecular structures, supporting its suitability for practical structure-based drug design.

\subsection{\textbf{3.4 Ablation Studies}}

\subsubsection{Effect of the MSCoD Interaction Mechanism on Target-specific Molecule Generation}
We conducted comprehensive ablation experiments to study the impact of different components in the MSCoD method on target-specific molecule generation capability. The experimental settings include four configurations: (1) Baseline represents the baseline model without applying the MSCoD method for comparison; (2) MSIB incorporates the Multi-Scale Information Bottleneck module for feature extraction and optimization through residual connections; (3) MHCA introduces the Multi-Head Cooperative Attention mechanism to capture interaction features between proteins and ligands; (4) MSCoD combines both MSIB and MHCA modules for joint feature extraction and multi-level information interaction. The experimental results are shown in Table~\ref{table:MSCoD-Interaction-Mechanism}. 

We observe that adding the MSIB module improves the mean Vina Score and median Vina Min values, demonstrating enhanced binding affinity prediction. Meanwhile, adding the MHCA module improves the median Vina Score, indicating more stable binding affinity prediction across different test cases. Furthermore, the combination of these two modules effectively improves QED values, indicating better drug-likeness properties in generated molecules.

\subsubsection{Effect of the Number of Attention Heads in the MHCA on Target-specific Molecule Generation} 
The number of attention heads in MHCA is a critical hyperparameter that significantly affects model performance and computational efficiency. Increasing the number of attention heads enables the model to capture interaction information between proteins and ligands from multiple perspectives simultaneously. This allows for more sophisticated modeling of complex binding interactions. When the number of attention heads is small, the model may not adequately capture the complex multi-level interactions between proteins and ligands. This leads to decreased quality of generated molecules, especially when generating molecules with strong target affinity and specific binding requirements.To verify the impact of the number of heads on model performance, we conducted comprehensive ablation experiments under multiple head settings including 4, 8, and 16 heads respectively. The experimental results are shown in Table~\ref{table:attention-heads}. 

We found that with 8 heads, the model achieves the best overall performance across all evaluation metrics with optimal balance. With 4 heads, the model's capture capability is significantly limited, resulting in suboptimal binding affinity predictions and reduced molecular quality. When the number of heads increases to 16, although the model's expressive capability improves theoretically, training complexity also increases significantly. This leads to decreased model performance under the original training schedule due to overfitting and increased computational overhead. Therefore, we recommend using the 8-head setting as the optimal balance point between performance and computational efficiency for practical applications.

\begin{figure*}[htbp]
    \centering
    \includegraphics[width=0.95\textwidth]{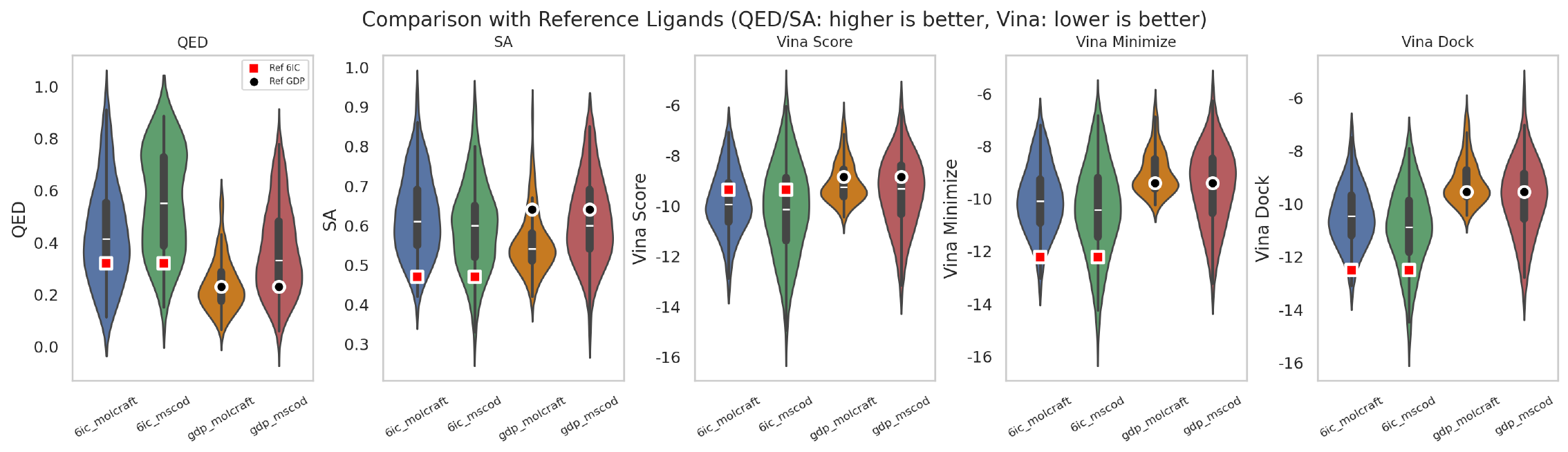}
    \caption{Property distributions (QED, SA, and Vina Score) for the 100 molecules generated by MSCoD for the two distinct binding pockets of the KRAS G12D (7XKJ) target, defined by the co-crystallized ligands GDP and 6IC, respectively.}
    \label{fig:7xkj}
\end{figure*}

\begin{table*}[htbp]
\centering
\scriptsize
\setlength{\tabcolsep}{4pt}
\begin{tabular*}{\textwidth}{@{\extracolsep{\fill}}c|ccccc|ccccc@{\extracolsep{\fill}}}
\toprule
\multicolumn{1}{c|}{} & \multicolumn{5}{c|}{Davis} & \multicolumn{5}{c}{Kiba} \\
\midrule
Methods & RMSE(↓) & MSE(↓) & Pearson(↑) & Spearman(↑) & CI(↑) & RMSE(↓) & MSE(↓) & Pearson(↑) & Spearman(↑) & CI(↑) \\
\midrule
GIN & 0.505 & 0.255 & 0.826 & 0.692 & 0.884 & 0.411 & 0.169 & 0.868 & 0.857 & 0.871 \\
GIN (MSCoD) & \textbf{0.492} & \textbf{0.242} & \textbf{0.837} & \textbf{0.698} & \textbf{0.888} & \textbf{0.402} & \textbf{0.162} & \textbf{0.875} & \textbf{0.873} & \textbf{0.888} \\
\midrule
GAT & 0.527 & 0.277 & 0.811 & \textbf{0.682} & \textbf{0.877} & 0.457 & 0.208 & 0.833 & 0.827 & 0.849 \\
GAT (MSCoD) & \textbf{0.500} & \textbf{0.250} & \textbf{0.831} & 0.673 & 0.874 & \textbf{0.425} & \textbf{0.181} & \textbf{0.857} & \textbf{0.851} & \textbf{0.873} \\
\midrule
GCN & 0.544 & 0.296 & 0.798 & 0.658 & 0.863 & 0.445 & 0.198 & 0.842 & 0.828 & 0.849 \\
GCN (MSCoD) & \textbf{0.512} & \textbf{0.262} & \textbf{0.821} & \textbf{0.672} & \textbf{0.874} & \textbf{0.395} & \textbf{0.156} & \textbf{0.879} & \textbf{0.875} & \textbf{0.891} \\
\midrule
GAT\_GCN & 0.532 & 0.282 & 0.806 & 0.676 & 0.874 & 0.408 & 0.167 & 0.869 & 0.861 & 0.873 \\
GAT\_GCN (MSCoD) & \textbf{0.508} & \textbf{0.258} & \textbf{0.827} & \textbf{0.695} & \textbf{0.887} & \textbf{0.387} & \textbf{0.150} & \textbf{0.884} & \textbf{0.878} & \textbf{0.892} \\
\botrule
\end{tabular*}
\caption{Drug-target affinity (DTA) prediction performance on the Davis and Kiba datasets. The methods are GraphDTA variants, where "(MSCoD)" indicates the baseline has been augmented with the MSCoD block (MSIB and MHCA). Boldface highlights the superior value in each baseline/augmented pair.}
\label{table:davis-kiba}
\end{table*}

\subsection{\textbf{3.5 Case Study: Generation for the Challenging 7XKJ Target}}

To evaluate the practical applicability of MSCoD, we performed conditional molecular generation experiments on the 7XKJ protein binding pocket, which corresponds to the clinically relevant KRAS G12D mutant. This target is known for its challenging binding environment, including a dynamic, shallow pocket and altered electrostatics due to the G12D mutation\citep{mao2022kras, akkapeddi2023exploring}. The 7XKJ structure contains two chemically distinct co-crystallized ligands (GDP and 6IC), providing a robust benchmark for generative model evaluation. We designed two experimental scenarios by defining separate binding pockets around each ligand using a radius-based protocol. For each scenario, MSCoD generated 100 molecules using only the protein structure as input, ensuring unbiased exploration. Performance was compared with MolCRAFT under identical conditions.

MSCoD demonstrated superior performance across all evaluation metrics, generating diverse, high-quality molecules suitable for the challenging KRAS G12D binding environment. The co-crystallized ligands establish performance benchmarks: GDP exhibits a Vina Score of -8.84 kcal/mol and QED of 0.229, while 6IC achieves -10.137 kcal/mol and 0.319, respectively. In the GDP-defined pocket scenario, MSCoD generated molecules with both enhanced binding affinity (-9.468 vs. -8.84 kcal/mol, a 7.1\% improvement) and significantly improved drug-likeness (QED: 0.369 vs. 0.229, a 61.1\% enhancement) compared to the reference GDP ligand. It also outperformed MolCRAFT with stronger Vina Scores (-9.468 vs. -9.032) and higher QED (0.369 vs. 0.233). In the 6IC-defined pocket scenario, MSCoD maintained an equivalent top-tier binding affinity (-10.137 kcal/mol) while delivering substantial improvements in drug-likeness (QED: 0.559 vs. 0.319, a 75.2\% enhancement) and outperforming MolCRAFT (QED: 0.559 vs. 0.422). These results demonstrate that MSCoD not only matches or exceeds the binding performance of proven co-crystallized ligands but also generates molecules with superior pharmaceutical properties, highlighting its potential for challenging oncogenic targets.

\subsection{\textbf{3.6 Transferability Study: Enhancing DTA Prediction with MSIB and MHCA}}
To evaluate the transferability of our proposed components, we integrated the MSIB and MHCA modules as a single, combined block—hereafter referred to as the MSCoD block—into four GraphDTA variants (GIN, GAT, GCN, and GAT\_GCN) for drug--target affinity (DTA) prediction~\citep{nguyen2021graphdta}. The experiments were strictly controlled to isolate the effect of this block: baseline architectures and all original hyperparameters remained unchanged. All models were trained under an identical protocol on the Davis and Kiba datasets, with results averaged over three independent runs (Table~\ref{table:davis-kiba}).

To assess performance on the DTA task, we employ a set of standard regression and ranking metrics. The Root Mean Square Error (RMSE) and Mean Squared Error (MSE) quantify prediction accuracy by measuring the deviation between predicted and actual binding affinity values; lower values indicate a more accurate model. For evaluating correlation, the Pearson coefficient measures the linear relationship, while the Spearman's rank coefficient assesses the monotonic relationship, indicating how well the model's predictions rank the compounds. Finally, the Concordance Index (CI) directly evaluates the probability that the model correctly ranks any two randomly chosen compounds. In drug discovery, strong performance in ranking metrics like Spearman and CI is particularly crucial, as it reflects a model's practical utility in correctly prioritizing potent compounds from a large candidate pool.

Across all encoder variants, augmenting GraphDTA with the MSCoD block led to consistent and measurable improvements. Notably, these gains were achieved without any further hyperparameter tuning, highlighting the "plug-and-play" utility of our approach. For example, the GAT\_GCN backbone benefited substantially: on Davis, RMSE decreased from 0.532 to 0.508 (a 4.5\% reduction) while CI increased from 0.874 to 0.887; on Kiba, RMSE decreased from 0.408 to 0.387 (a 5.1\% reduction) and CI rose from 0.873 to 0.892. Comparable gains were observed for the GIN, GAT, and GCN backbones, indicating that the performance enhancement is not specific to a single encoder design.

The magnitude of the gains is modest but consistent (typical RMSE reductions of ~0.02–0.05 and small increases in Pearson/Spearman/CI), which suggests that MSIB improves hierarchical feature extraction while MHCA strengthens asymmetric protein→ligand interaction modeling. Under an identical training protocol, these results support the modules' practical transferability: inserting MSIB and adding MHCA to diverse GraphDTA backbones enhances affinity-prediction accuracy and ranking, demonstrating the utility of MSCoD components for cross-task applications in structure-based drug discovery.

\section{\textbf{4 Conclusion}}\label{sec4}
In this work, we propose MSCoD, a novel generative framework that significantly advances the modeling of protein-ligand interactions by utilizing multi-scale information bottleneck modules and multi-head cooperative attention mechanism. MSCoD provides a robust and extensible framework for structure-based drug design that enables the precise generation of drug-like molecules with high binding affinity and favorable molecular properties. Experimental studies show that MSCoD achieves superior overall performance on structure-based molecular generation benchmarks. Case studies on challenging targets such as KRAS G12D further demonstrate its effectiveness for difficult protein targets, and the MSIB and MHCA modules also transfer effectively to drug–target affinity prediction, enhancing predictive performance on standard datasets. In summary, this work has the potential to expedite the drug discovery process.

However, some limitations remain, primarily due to computational constraints. Future work will focus on further enhancing the model's capacity to capture deeper and more complex protein-ligand interactions, for example by embedding more systematic attention mechanisms within multi-scale feature extraction.

\section{Acknowledgments}
The authors would like to thank the anonymous reviewers for their valuable suggestions that improved the paper.

\section{Conflict of interest}
None declared

\section{Funding}
Coming soon.

\bibliographystyle{apalike}
\bibliography{references}

\end{document}